# Predicting Preschoolers' Externalizing Problems with Mother-Child Interaction Dynamics and Deep Learning

Xi Chen[1,*], Yu Ji[2,3,*], Cong Xia[1], Wen Wu[1,3]

[1]Shanghai Key Laboratory of Mental Health and Psychological Crisis Intervention, School of Psychology and Cognitive Science, East China Normal University, Shanghai 200333, China

[2]School of Electronic and Electrical Engineering, Shanghai University of Engineering Science, Shanghai 201600, China

[3]School of Computer Science and Technology, East China Normal University, Shanghai 200333, China

*These authors contributed equally to this work

Correspondence: Xi Chen, Shanghai Key Laboratory of Mental Health and Psychological Crisis Intervention, School of Psychology and Cognitive Science, East China Normal University, Shanghai 200333, China, Email xchen@psy.ecnu.edu.cn; Wen Wu, Shanghai Key Laboratory of Mental Health and Psychological Crisis Intervention, School of Psychology and Cognitive Science, School of Computer Science and Technology, East China Normal University, Shanghai 200333, China, Email wwu@cc.ecnu.edu.cn

Abstract:

**Objective:** Predicting children's future levels of externalizing problems helps to identify children at risk and guide targeted prevention. Existing studies have shown that mothers providing support in response to children's dysregulation was associated with children's lower levels of externalizing problems. The current study aims to evaluate and improve the accuracy of predicting children's externalizing problems with mother-child interaction dynamics.

**Method:** This study used mother-child interaction dynamics during a challenging puzzle task to predict children's externalizing problems six months later ($N$=101, 46 boys, $M_{age}$=57.41 months, $SD$=6.58).

Performance of the Residual Dynamic Structural Equation Model (RDSEM) was compared with the Attention-based Sequential Behavior Interaction Modeling (ASBIM) model, developed using the deep learning techniques.

**Results:** The RDSEM revealed that children whose mothers provided more autonomy support after increases of child defeat had lower levels of externalizing problems. Five-fold cross-validation showed that the RDSEM had good prediction accuracy. The ASBIM model further improved prediction accuracy, especially after including child inhibitory control as a personalized individual feature.

**Conclusions:** The dynamic process of mother-child interaction provides important information for predicting children's externalizing problems, especially maternal autonomy supportive response to child defeat. The deep learning model is a useful tool to further improve prediction accuracy.

# Introduction

Preschoolers with symptoms of externalizing problems, such as aggression, oppositional behaviors, attention deficits, and hyperactivity, are at risk of a cascading process in which early symptoms sustain and contribute to negative developmental outcomes in the long run, such as poor social relationships, substance use, and unemployment.[1-3] Thus, it is important to identify the children who are likely to develop high levels of externalizing problems and provide prevention accordingly from early on. During the preschool years, children's ability to regulate their emotions and behaviors develops rapidly and serves as an important protective factor against externalizing problems.[4,5] A primary proximal context for the development of children's regulation ability is the dynamic process of parent-child interaction, in which both the parent and the child modulate their emotions and behaviors in response to each other's ongoing emotions and behaviors.[6-9] Mothers providing support in response to children's negative emotions and dysregulated behaviors may facilitate children's self-regulation and has been found to be significantly associated with children's lower levels of externalizing problems.[10-12] Yet, the accuracy of predicting children's externalizing problems with mother-child interaction dynamics has not been evaluated.

In this study, we aimed to predict preschoolers' externalizing problems six months later using maternal and child behaviors observed during a challenging puzzle game and evaluate the prediction accuracy using five-fold cross-validation. In addition, we compared the performance of the Attention-based Sequential Behavior Interaction Modeling (ASBIM) model, which we proposed using deep learning technique, with a Residual Dynamic Structural Equation Model (RDSEM),[13] which is an innovative model for time-series analysis. In addition, because children's inhibitory control (i.e., the ability to inhibit inappropriate behaviors and responses) has been found to be a protective factor against externalizing problems,[14-16] we also included children's inhibitory control in the ASBIM model and the RDSEM to further improve the prediction accuracy.

## *Mother-child Interaction Dynamics and Children's Externalizing Problems*

In coordinated parent-child mutual regulation, the parent and the child interact to fulfill the child's goals of maintaining harmonious internal status and engaging with the external environment.[9] On the one hand, the parent helps to repair disruptions of the child's ongoing activity, by providing support in response to the child's cues of dysregulation.[10,17] On the other hand, with parental support the child is able to change from a negative emotional status to a more positive emotional status and to pursue goal-directed engagement with people and objects.[10,18] Such a process is an opportunity for the child to practice effective regulation with parental support and may promote the child's self-regulation and reduce the child's externalizing problems in the long run.[19]

   Regarding implications of parent-child mutual regulation for children's externalizing problems, existing studies have mainly focused on the coercive process, that is, both the parent and the child responding to each other's coercive behaviors (i.e., parental intrusiveness and control, child noncompliance and defiance) with escalating coercion until the parent gives in, which contributes to children's conduct problems.[2,20] In support of the coercive process framework, mothers' negative emotions and harsh parenting in response to children's negative emotions and aversive behaviors have been found to predict children's more externalizing problems.[21,22] To our knowledge, only one study examined the dynamics between parental support and children's negative affect as predictors of children's externalizing problems. This study found that during a conflict discussion task, the likelihood of mothers showing supportive responses to children's negative affect and the likelihood of children transiting out of negative emotion in response to maternal support was higher for typically-developing 8-to-12-year-old children than for children with externalizing problems.[10]

During the preschool period, parent-child coordinated mutual regulation may be especially apparent in challenging problem-solving tasks, in which the parent and the child work together to facilitate the child's regulation of frustration and engagement with the task.[17,23] Parental autonomy support (i.e., respecting children's perspectives, giving children choices, and encouraging children to take ownership of their work) may serve as an important approach to support preschoolers' self-regulation in challenging problem-solving tasks.[18,24,25] The likelihood that mothers transited from less autonomy-supportive behaviors to more autonomy-supportive behaviors, when 3.5-year-old children were autonomously engaged with the task, and the likelihood that mothers responded to children's compliance with teaching (i.e., explanation and pedagogical questions) rather than directives, predicted children's lower levels of externalizing and internalizing problems four months later.[24] Implications of the dynamics between maternal autonomy support and preschoolers' dysregulated behaviors during challenging puzzle remain to be examined.

## *Children's Inhibitory Control and Externalizing Problems*

Children's inhibitory control is a key aspect of temperament related to self-regulation,[26] which develops rapidly during the preschool period.[27-29] Children with higher inhibitory control are able to regulate their emotions and behaviors more consciously and flexibly, and suppress socially undesirable behaviors.[29,30] Thus, children with higher inhibitory control have been found to show lower levels of externalizing problems.[14-16]

Besides the direct link between children's inhibitory control and externalizing problems, children with different levels of inhibitory control may also respond to and be affected by parenting behaviors in different ways. Children with lower inhibitory control have more difficulty regulating their emotions and behaviors and may need parental support to a greater extent. In support of this hypothesis, a few studies found that only for children with low inhibitory control, maternal warmth predicted fewer externalizing problems,[31] and maternal physical punishment and negative control predicted more externalizing problems.[32,33] In contrast, the findings of another study supported the goodness of fit framework. That is, for children with low inhibitory control, maternal autonomy support predicted more externalizing problems and maternal negative control predicted fewer externalizing problems, whereas for children with high inhibitory control, maternal autonomy support predicted fewer externalizing problems and maternal negative control predicted more externalizing problems.[14] Despite differences in specific patterns of the interaction, these findings all suggest that considering child inhibitory control together with parent-child interaction dynamics may contribute to more accurate prediction of children's externalizing problems.

## *Predicting Child Externalizing Problems Using RDSEM and ASBIM Model*

The existing studies about predictors of children's externalizing problems have focused on testing whether the variance of children's externalizing problems can be explained by the interested predictors, but did not evaluate the accuracy of prediction using certain predictors or models. The latter is of important practical implications, because it helps to identify the children who are at greater risk in order to inform early prevention.

One of the latest method to model the mutual regulation process of parent-child interaction is the RDSEM, which is a combination of multi-level modeling and structural equation modeling for time series analysis.[13] In the RDSEM, at the within-person level, the association between fluctuations of maternal behaviors in a given moment with fluctuations of child behaviors in the next moment (i.e., mother-to-child lagged association) and the association between fluctuations of child behaviors in a given moment with fluctuations of maternal behaviors in the next moment (i.e., child-to-mother lagged association) can be estimated in the same model to reflect the mutual regulation process. In addition, autoregressions between the residuals of maternal and child behaviors and the covariance between maternal and child behaviors at the same moment can also be modeled. At the between-person level, children's externalizing problems at Time 2 can be predicted using the mother-to-child and child-to-mother lagged associations, together with person means of maternal and child behaviors, demographic variables and child externalizing problems at Time 1. Children's inhibitory control can also act as a predictor at the between-person level, but the RDSEM cannot estimate the interactions between inhibitory control and the latent variables of mother-to-child or child-to-mother lagged associations or the interactions between inhibitory control and the person means of maternal and child behaviors.

Although the mother-to-child and child-to-mother associations can be used to estimate the mutual regulation process in mother-child interaction, they may not be able to fully depict the complex dynamics of interaction. For example, the mother and the child may not only be affected by what each other did in the previous moment but may also be affected by the accumulation of certain behaviors from the beginning of the interaction; the mother-to-child and child-to-mother associations may not be constant throughout the interaction; maternal and child behaviors at different time during the interaction may have different implications for child adjustment. Thus, a more data driven approach may be able to capture the complex dynamics that are useful for predicting children's externalizing problems. Deep learning model could model the behavior interaction at different time and extract high-order interaction features,[35] enabling a more effective and fine-grained analysis of the behavior sequences of mother and child compared to RDSEM. Another advantage of the deep learning model is that demographic variables, person means of maternal and child behaviors, as well as children's inhibitory control can be included as personalized individual features, thus interaction between these variables and the dynamics of mother-child interaction can be taken into consideration when predicting children's externalizing problems, which may further improve prediction accuracy.

A few studies have begun to develop various computational tools to predict variables with complex mechanisms.[36,37] For example, Paolucci et al. (2023) used a machine learning model to distinguish individuals with autism spectrum disorders from controls based on ratings for home videos of children doing everyday activities.[38] Yan et al. (2023) obtained a dataset of students' demographics, physical and mental health, peer and family relationships, bullying experience and combined two machine learning algorithms to predict bullying victims.[39] Waheed et al. (2020) applied Artificial Neural Networks to improve the prediction accuracy of students' academic performance from a big longitudinal dataset.[40] To our knowledge, there haven't been any studies using deep learning techniques and the dynamic data of parent-child interaction to predict future child developmental outcomes. Therefore, we propose the Attention-based Sequential Behavior Interaction Modeling (ASBIM) model to enhance the prediction accuracy of children's externalizing problems. By incorporating mother and child features and analyzing the mother-child interaction sequence, the ASBIM model can effectively predict the child's externalizing problems.

As a commonly used method for evaluating prediction accuracy in machine learning and deep learning, five-fold cross-validation (where *k* is typically set to five for small datasets) is employed in our work to validate the performance of both RDSEM and our ASBIM model. Specifically, in *k*-fold cross-validation, the dataset is initially partitioned randomly into *k* distinct folds, each containing a roughly equal number of samples. Subsequently, each fold takes turns serving as the test set for the model trained on the remaining *k*-1 folds.[34]

## *The Current Study*

In the current study, we aimed to examine the accuracy of predicting children's externalizing problems with maternal autonomy support and child defeat (i.e., expression of frustration, incapacity to complete the task or giving up) during a challenging puzzle task using a more theory-driven RDSEM and a more data-driven ASBIM model. We also examined to what extent adding children's inhibitory control in the models could improve prediction accuracy. Findings of this study would help to predict children's externalizing problems and inform early prevention.

# Method

## *Participants*

The participants were drawn from a larger study examining mother-child interaction and children's socioemotional functioning during the preschool period in China (see[41]). Participants were recruited in three preschools in a small city in Middle China in 2018. The sample size was determined referring to others observational studies of the dynamic process of mother-child interaction in challenging problem-solving tasks during the preschool period.[23,24] A recent study using the RDSEM model reported a similar sample size.[42] Of the 113 families participated, 101 families (46 boys, $M_{age}$=57.41 months, *SD*=6.58) were included in the analyses. Reasons for exclusion were child refusing to work on or quitting the puzzle task within the first minute (*n*=4), child not understanding how to play the puzzle (*n*=1), mother did not adhere to task instructions (e.g., looked at child's puzzle, *n*=3), technical errors (*n*=2), and child significantly younger (38 months old) or older (77 months old) than other children (*n*=2). In the families included, on average, the mothers were 31.93 years of age (*SD*=3.43, Range=23-42 years) and received 14.64 years of education (*SD*=2.48, Range=10-20 years), and the fathers were 33.18 years of age (*SD*=3.63, Range=23-47 years) and received 14.47 years of education (*SD*=2.44, Range=10-20 years). Annual income of the families ranged from 10,000 yuan to 500,000 yuan (about $4,398 to $73,301), with a median of 100,000 yuan (about $14,660). The participants excluded did not differ from those included on these demographic variables or the study variables. The parents provided written consent and the children provided oral consent for participation in this study. The study was approved by the Institutional Review Board at East China Normal University (protocol no. HR2-1118-2020). This study was not preregistered. Study materials and analysis codes are available upon request. Data of this study can be obtained at https://dx.doi.org/10.21227/cyyj-7378.

## *Procedure*

This study was longitudinal with two time points, which were six months apart. At T1, the mother-child dyads participated in a 45-min research visit in a quiet room in the child's preschool and their interaction was video recorded. At both T1 and T2 mothers and fathers reported child externalizing problems. At T1 mothers and fathers reported child inhibitory control.

In this report, we focused on mother-child interaction in the puzzle tasks, which followed a 10-min mother-child play session. During each of the two puzzle tasks, mothers and children were presented with a 14- piece Tangram® puzzles (i.e., dog or lion), where the level of challenge was beyond what a preschooler was capable of completing independently. Mothers and children were seated across from each other at a child-sized table, with a barrier positioned in the middle of the table. During one puzzle task, the barrier was set low so that the mother could see her child but not the puzzle board or pieces; during the other puzzle task, the barrier was set high so that the mother could not see either her child or the puzzle. The order of the puzzle (dog, lion) and barrier type (low, high) were counterbalanced across participants. The child was given the puzzle board and pieces, and the mother was given the solution to the puzzle. The mother and child were instructed to try and complete the puzzle in five minutes and to communicate only using words. A large countdown timer was visible to both the mother and the child and signaled the end of the task.

The puzzle tasks were designed specifically to address an objective of the larger study on maternal speech prosody and children's behavioral and physiological regulation. Given the focus on mother-child interaction in this report, we only examined data from the low barrier condition, where mother-child dyads could use visual cues (e.g., expression, gaze, gestures) during their interaction, and we excluded examination of the high barrier condition.

## *Measures*

## **Observational Coding of Maternal and Child Behaviors**

Maternal and child behaviors during the puzzle task were rated by different teams of coders (three coders on each team), who were Chinese undergraduate students majored in psychology. Behaviors were rated on 4-point scales (0=*not at all characteristic*, 3=*highly characteristic*) in 15-sec intervals, and the ratings captured frequency, duration and/or intensity of behaviors.

*Maternal autonomy support* reflects the degree to which the mother acted in a way that recognized and respected the child's individuality, motives, and perspective. This code was modified from existing studies (see[18,43,44]). Maternal autonomy support was demonstrated through positive and engaged response to the child's bids for attention or help (e.g., giving a hint or explanation when the child asks for help), interest in and encouragement of what the child was doing and saying (e.g., verbal acknowledgement of the child's statement), provision of opportunities for the child's independence or input (e.g., asking a pedagogical question), positive acknowledgement or encouragement of the child's effort (e.g., "Good job!", "Have a try.") and emotional support when necessary (e.g., "I know it is difficult. Don't worry."). It was difficult for coders to reach acceptable levels of reliability in training due to complexity of this behavior. Thus, maternal autonomy support for all dyads were double coded. Discrepancies of interval ratings greater than one were discussed to reach consensus and other ratings were averaged between the two coders. Because the two coders' ratings were averaged to get the final ratings for each interval, intraclass correlations (ICCs) for averaged measures[45] were calculated to assess interrater reliability (ICC=.75).

*Child defeat* was captured by two codes, *behavioral defeat* and *negative affect*. *Behavioral defeat* captures the child's uncertainty, reluctance, or withdrawal from the task. Behavioral defeat was demonstrated by (a) verbal statements indicating the child was struggling with the puzzle task (e.g., "I cannot get this piece to fit!"), giving up, or feeling a lack of confidence in one's ability to solve the puzzle (e.g., "This is too hard!") and (b) non-verbal signs indicating withdrawal from or giving up on the task (e.g., pushing the puzzle away). *Negative affect* captures the child's displays of sadness or anger, such as frowning, glaring, pouting, whining, and harsh tone of voice (modified from[46]). For the child behaviors, the coders reached acceptable levels of reliability in training. Thus, following the typical approach for assessing interobserver reliability, 20% of the tapes were double coded for child behaviors, and reliability was assessed throughout the official coding process. For child codes, therefore, only one coder's ratings were used as the final ratings, and ICCs for single measures[45] were calculated to assess interrater reliability (ICC=.71 and .79, for behavioral defeat and negative affect, respectively). Ratings of child behavioral defeat and negative affect were positively correlated at the within-person level ($r$=.63, 95% confidence interval [.54, .70]) and were averaged in each interval to obtain scores of child defeat.

## Parental Reports of Children's Externalizing Problems

At T1, all 101 mothers and 87 fathers completed the Strengths and Difficulties Questionnaire (SDQ[47]) and at T2, 95 mothers and 87 fathers completed the questionnaire. SDQ is a widely used questionnaire to assess children's behavioral adjustment. The copyrighted Chinese version for parents of 4- to 17-year-olds was obtained from https://www.sdqinfo.org. There is support for good reliability and convergent validity of SDQ in Chinese population.[48] In this report, we focused on parental reports of externalizing problems (10 items, e.g., "Restless, overactive, cannot stay still for long", "Often has temper tantrums"). Parents rated each item on a 3-point scale ranging from 0 (not true) to 2 (certainly true) based on their children's behaviors over the last six months, and the ratings (reverse scored when appropriate) were averaged across the items. Internal reliability was acceptable to good (Cronbach α=.69-.70). Mother and father reports were positively correlated ($r$=.52 and .63, $ps$<.001 at T1 and T2, respectively) and averaged.

## Parental Reports of Children's Inhibitory Control

The short form of Children's Behavior Questionnaire (CBQ-SF[49]) was used to measure children's temperament. Previous study showed that CBQ-SF had good internal reliability in Chinese samples.[50] In this report, we examined parental reports on child inhibitory control, which was measured by 6 items (e.g., "Is good at following instructions") on a 7-point Likert scale (1=extremely untrue of your child, 7=extremely true of your child) based on child behaviors over the last six months. If parents have not observed their child reactions in certain situations, they can select the *not applicable* option. One item was reverse scored and the mean score of the six items was calculated. Internal reliability of the inhibitory control subscale was good (Cronbach α=.79 and .74 for mother and father reports, respectively). Father-and mother-reported scores were significantly correlated ($r=.51$, $p<.001$) and averaged to create composite scores of inhibitory control.

## *Data Analytic Strategies*

### Data preprocessing and missing data

Scores for child defeat were heavily skewed toward 0 (0 in 78% of the intervals during mother-child interaction), and these ratings were recoded into a binary variable (i.e., 0=*not present*, 1=*present*). Although the frequency of child defeat was low, only 10.9% of the children did not show any defeat during the task, and 76.2% of the children showed two or more instances of defeat.

In 0.4% and 0.3% of the intervals, maternal autonomy support and child defeat were missing because the coders could not hear the mother's or the child's speech due to video quality or because the mother-child dyad were talking about topics irrelevant to the puzzle game. Of the 101 families included in the analyses, 5 were missing parental reports of child behavioral adjustment at T2. We first fitted the RDSEM using the full sample to reveal the relations between the study variables. At this stage, Bayesian estimation was used to handle missing data. Next, we evaluated the prediction accuracy of RDSEM using the five-fold cross-validation. At this stage, 10 datasets were imputed. We conducted five-fold cross-validation for each imputed dataset. Next, we examined the predication accuracy of ASBIM using five-fold cross-validation for each imputed dataset.

### RDSEM

As preliminary steps, (a) child gender, age, and maternal years of education were examined as potential covariates by testing their associations with externalizing problems at T2, and (b) systematic changes in maternal and child behaviors during the puzzle task were examined by testing the effect of time (i.e., intervals) on maternal autonomy support and child defeat using RDSEMs. The significant covariates and effects were retained in the main model described next.

In the main model, at the within-person level, paths were estimated from maternal autonomy support in the current 15-sec interval (*t*) to child defeat in the next interval (*t*+1, i.e., mother-to-child lagged association), and from child defeat in the current 15-sec interval (*t*) to maternal autonomy support in the next interval (*t*+1, i.e., child-to-mother lagged association). The covariance between maternal and child behaviors within the same 15-sec interval and the autoregressive paths for residuals of maternal and child behaviors were also estimated.

At the between-person level, the bi-directional lagged associations between maternal autonomy and child defeat and the person means of maternal autonomy support and child defeat during the puzzle task were examined as predictors of child behavioral adjustment at T2, while controlling for child adjustment at T1. Variances of the bi-directional lagged associations and the autoregressive paths, and their covariances with each other and with the between-person predictors were also estimated. The models were estimated both with and without child inhibitory control as a between-person predictor.

The five-fold cross-validation. We first fit the RDSEM using the full sample to reveal the relations between the study variables and then evaluated the prediction accuracy using the five-fold cross-validation. For the model without child inhibitory control as a predictor, for each training dataset, the RDSEM was fitted to obtain estimations of the intercept for externalizing problem at T2, the coefficients of the predictors (i.e., the mother-to-child lagged association, the child-to-mother lagged association, the person mean of maternal autonomy support, the person mean of child defeat) and the coefficients of the covariates (i.e., child externalizing problems at T1 and child gender). For each testing dataset, a simplified RDSEM was fitted to estimate the mother-to-child and child-to-mother lagged associations and person means of maternal autonomy support and child defeat for each mother-child dyad. In the simplified RDSEM, only the within-person effects (i.e., mother-to-child lagged association, child-to-mother lagged association, autoregressions of maternal and child behaviors, time effects on maternal and child behaviors), and their covariances with each other, with person means of maternal and child behaviors and with child externalizing problems at T1 were estimated. For each mother-child dyad in the testing dataset, the predicted value of children's externalizing problems at T2 was calculated using the estimated values of mother-to-child and child-to-mother lagged associations, person means of maternal autonomy support and child defeat, the observed values of child externalizing behaviors at T1, child gender and the intercept and coefficients estimated in the corresponding training dataset. The average squared difference (Mean Square Error, MSE) and the average correlation between the predicted values and the observed values of children's externalizing problems at T2 across the five testing datasets were calculated to evaluate the prediction accuracy. For the model with child inhibitory control as a between-person predictor, the procedure was the same, except that for each training dataset the coefficient of child inhibitory control was also estimated, and that for each testing dataset the observed values of child inhibitory control were also used to predict child externalizing problems at T2.

## ASBIM

Before we elaborate the details of our ASBIM model, we give the formalization first. Concretely, for maternal $m \in M$ and child $c \in C$, $S^m=\{s^m_1, s^m_2, ..., s^m_n\}$ and $S^c=\{s^c_1, s^c_2, ..., s^c_n\}$ represent the behavior sequence of maternal $m$ and child $c$, while $s^m_i$ and $s^c_i$ stand for the autonomy support of maternal $m$ and the defeat of child $c$ in the $i$-th 15-sec interval respectively. In addition, maternal $m$ and child $c$ have the same length of the behavior sequence, while $n$ denotes the length of the behavior sequence. Besides, $g^c$ represents the gender of child $c$. In addition, $e^c_{T1}$ and $e^c_{T2}$ denote the externalizing problems of child $c$ at T1 and T2 respectively. Then, the regression task of children' externalizing problems prediction is defined as: Given $S^m$, $S^c$, $g^c$, and $e^c_{T1}$, we want to predict $e^c_{T2}$.

Figure 1 depicts the whole framework of our ASBIM model, which includes four modules. Concretely, in *individual feature modeling module*, ASBIM model acquires the embedding vectors of mother and child features for modeling the behavior representations of mother and child in a personalized manner subsequently. Then, in *behavior interaction representation modeling module*, ASBIM model generates the mother-child behavior interaction representations based on the individual feature representations. After that, in *behavior interaction pattern modeling module*, ASBIM model extracts the mother-child behavior interaction pattern from the corresponding sequence of behavior interaction representations. Finally, in *prediction module*, ASBIM model adopts the mother-child behavior interaction pattern to predict the child's externalizing problem at T2.

Specifically, in *individual feature modeling module*, we adopt look-up embedding technique[51] to obtain the embedding vectors of mother and child features. To be specific, the individual features utilized in this module not only include the gender of child c and the externalizing problems of child c at T1 but also incorporate the person mean of maternal *m*'s autonomy support and child *c*'s defeat. By including these personalized features, our ASBIM model can account for unique interaction patterns between each mother and child, thereby enabling a more accurate and personalized understanding the child's externalizing problem. Concretely, we refer to $p^m_c=\{p^{mc}_1=\overline{s^m}, p^{mc}_2=\overline{s^c}, p^{mc}_3=e^c_{T1}, p^{mc}_4=g^c\}$ as the individual feature set, where $\overline{s^m} = \dfrac{\sum_{i=1}^{n} s_i^m}{n}$ and $\overline{s^c} = \dfrac{\sum_{i=1}^{n} s_i^c}{n}$ denote for the person mean of corresponding maternal autonomy support and child defeat respectively. We then adopt look-up embedding technique to obtain the embedding vector of each mother/child feature in $p^{mc}$. In particular, the individual features have two types: a categorical type with a finite set of distinct values (i.e., $g^c$) and a numerical type characterized by continuous values (i.e., $\overline{s^m}$, $\overline{s^c}$, and $e^c_{T1}$). We formulate the embedding matrix or the look-up table of the $i$-th individual feature as:

$$\varphi_i \in \begin{cases} \mathbb{R}^{l_i \times q} & \text{the } i-\text{th individual feature is categorical type} \\ \mathbb{R}^q & \text{the } i-\text{th individual feature is numerical type} \end{cases} \quad (1)$$

where $l_i$ stands for the number of distinct values in the $i$-th categorical type and $q$ denotes the dimension of the embedding vectors. Then, we obtain the embedding vector of $\mathbf{p}^{mc}_i$ by:

$$\mathbf{p}^{mc}_i \in \begin{cases} \varphi_i[p_i^{mc}] & \text{the } i-\text{th individual feature is categorical type} \\ p_i^{mc} \times \varphi_i & \text{the } i-\text{th individual feature is numerical type} \end{cases} \quad (2)$$

where $\varphi_i[p_i^{mc}]$ represent the $p^{mc}_i$-th row of $\varphi_i$.

Finally, we obtain the representation of the mother and child features by a simple and efficient average operation:

$$\mathbf{p}^{mc} = avg\left(\mathbf{p}^{mc}_i\right) \quad (3)$$

where $avg(\cdot)$ represents the averaging operation across all input vectors.

In *behavior interaction representation modeling module*, we adopt the representation of the mother and child features to obtain the personalized mother-child behavior interaction representations. Concretely, we first utilize the individual feature representation $\mathbf{p}^{mc}$ to get the representation of each mother/child behavior in $S^m$ or $S^c$. To be specific, the mother behavior representation $\mathbf{s}^m_i$ corresponding to the mother behavior $s^m_i$ is obtained via:

$$\mathbf{s}^m_i = (s_i^m - \overline{s^m}) \times relu(\mathbf{W}_1 \mathbf{p}^{mc} + \mathbf{b}_1) \quad (4)$$

where $relu(x)=max(0, x)$ is activation function. $\mathbf{W}_1$ and $\mathbf{b}_1$ are weights and biases respectively. $s_i^m - \overline{s^m}$ represents fluctuations of maternal behavior from moment to moment after taking out the person mean. Similarly, the child behavior representation $\mathbf{s}^c_i$ corresponding to the child behavior $s^c_i$ is computed as:

$$\mathbf{s}^c_i = (s_i^c - \overline{s^c}) \times relu(\mathbf{W}_2 \mathbf{p}^{mc} + \mathbf{b}_2) \quad (5)$$

where $\mathbf{W}_2$ and $\mathbf{b}_2$ are weights and biases respectively.

Then, the mother-child behavior interaction representation $\mathbf{s}^{mc}_i$ is obtained by a weighted summation of $\mathbf{s}^m_i$ and $\mathbf{s}^c_i$:

$$\mathbf{s}^{mc}_i = \gamma \mathbf{s}^m_i + (1-\gamma)\mathbf{s}^c_i \quad (6)$$

where γ is trainable weight for measuring the importance of maternal autonomy support and child defeat in the regression task of children' externalizing problems prediction.

In *behavior interaction pattern modeling module*, we adopt the self-attention mechanism[52] to model the behavior interaction sequence of mother and child and extract the mother-child behavior interaction pattern $\mathbf{S}^{mc}$. Although certain existing approaches utilize such as Long-Short Term Memory (LSTM) to model behavior sequences,[53] they typically face the challenge of insufficient explainability. However, leveraging the self-attention mechanism enables us to identify the significance of each mother-child behavior interaction in representing the mother-child interaction pattern through attention weights. Concretely, $\mathbf{S}^{mc}$ is acquired by:

$$\mathbf{S}^{mc} = \sum_{i=1}^{n} \alpha_i f_1\left(\mathbf{s}^{mc}_i\right) \quad (7)$$

$$\alpha_i = \frac{\exp\left(\frac{\langle f_2(\mathbf{s}^{mc}_i), f_3(\mathbf{s}^{mc}_i)\rangle}{\sqrt{h}}\right)}{\sum_{i=1}^{n}\exp\left(\frac{\langle f_2(\mathbf{s}^{mc}_i), f_3(\mathbf{s}^{mc}_i)\rangle}{\sqrt{h}}\right)} \quad (8)$$

where $\alpha_i$ is the attention weight of $\mathbf{s}^{mc}_i$. $\langle \cdot, \cdot \rangle$ represents the inner product, while $f_1(\cdot)$, $f_2(\cdot)$, and $f_3(\cdot)$ stand for the dense layer to project the mother-child behavior interaction representation $\mathbf{s}^{mc}_i$ to one new vector representation.

In *prediction module*, based on $e^c_{T1}$, we adopt the mother-child behavior interaction pattern $\mathbf{S}^{mc}$ to obtain the final prediction result $\hat{y}^c$:

$$\hat{y}^c = \mathbf{W}_3 \mathbf{S}^{mc} + \mathbf{b}_3 + e^c_{T1} \quad (9)$$

where $\mathbf{W}_3$ and $\mathbf{b}_3$ are weights and biases respectively.

The loss function for optimization is defined as mean square error between the gold truth and our ASBIM model's prediction when training:

$$L = \sum_{c \in C}\left(\hat{y}^c - e^c_{T2}\right)^2 \quad (10)$$

where *C* represents the child set.

For the implementation details, we adopt Adam[54] as our ASBIM model's optimizer to update parameters when training, while the learning rate is set to 1e-3. Besides, the batch size is set to the size of the training set as the total dataset is small (i.e., 101 samples). Furthermore, individual mother-child interaction sequences can vary in length. Given that all sequences are no longer than 20, we establish the maximum sequence length as 20 (padding when sequence length is less than 20, and truncating when sequence length is greater than 20) to streamline model processing. Consequently, the ASBIM model can handle each mother-child interaction sequence according to its specific length, up to the established maximum threshold. Moreover, the dimension of the embedding vectors is set to 50 (i.e., $q$=50). Given the small size of the total dataset (101 samples), we employ L2 regularization to boost the generalization capability of our model and mitigate the risk of overfitting. Additionally, we opt for a batch size equivalent to the dataset size, enabling full-batch gradient descent. This choice helps stabilize the training process by minimizing gradient noise.

Furthermore, considering the effectiveness of the child's inhibitory control, we also implement a variation (i.e., ASBIM+D) of our ASBIM model, which incorporates the child's inhibitory control as an individual feature. To be specific, ASBIM+D adopts $p^{mc}=\{p^{mc}_1=\overline{s^m}, p^{mc}_2=\overline{s^c}, p^{mc}_3=e^c_{T1}, p^{mc}_4=g^c, p^{mc}_5=d^c\}$ to model the individual features, while $d^c$ represents the inhibitory control of child $c$.

# Results

## *Descriptive Statistics and Bivariate Correlations*

Descriptive statistics for the study variables at the between-person level (i.e., person means of maternal autonomy support and child defeat at T1, child externalizing problems at T1 and T2, child inhibitory control at T1), their bivariate correlations (between-person correlations are below the diagonal and within-person correlations are above the diagonal), and ICCs (i.e., the proportion of between-person variance in total variance) of maternal autonomy support and child defeat are presented in Table 1. The ICCs show that both maternal autonomy support and child defeat varied more within each mother-child dyad than between different families. Each mother's mean levels of autonomy support throughout the puzzle task was not correlated with their children's mean levels of defeat throughout the task, or children's externalizing problems at T1 and T2. Yet, at the moment when the child expressed more defeat, the mother showed more autonomy support, indicating the importance to examine interaction dynamics within each mother-child dyad. On average, children with higher inhibitory control expressed less defeat throughout the task, and showed fewer externalizing problems at T1 and T2. Children's externalizing problems at T1 and T2 was positively correlated showing steability of children's externalizing problems across six months.

## *RDSEM*

### Preliminary analyses

Preliminary analyses showed that boys had higher levels of externalizing problems at T2 than girls ($t$=2.27, $p$=.025). Thus, child gender was included as a covariate in the main model. Child age ($r$=-.14, $p$=.184) and maternal education ($r$=-.06, $p$=.593) were not associated with children's externalizing problem at T2 and were not included in the main model.

The RDSEMs examining the effect of time (i.e. intervals) on maternal autonomy support and child defeat showed that maternal autonomy support (Estimate=.00, posterior $SD$=.01, 95% CI=[-.03, .02]) and child defeat (Estimate=.02, posterior $SD$=.02, 95% CI=[-.01, .05]) did not significantly change with time, whereas the corresponding random effects were significant (for maternal autonomy support: Estimate=.01, posterior $SD$=.00, 95% CI=[.01, .02]; for child defeat: Estimate=.02, posterior $SD$=.00, 95% CI=[.01, .03]). To account for individual differences in these systematic changes, fixed and random effects of time on maternal autonomy support and child defeat were estimated in the main model.

### The Main Model

The results of the main model using the full sample with and without child inhibitory control as a predictor are presented in Table 2. At the within-person level, higher maternal autonomy support at $t$ predicted higher child defeat at $t$+1. Child defeat at $t$ did not significantly predict maternal autonomy support at $t$+1. In addition, maternal autonomy support was positively associated with child defeat in the same interval. Autoregressive effects emerged for both maternal autonomy support and child defeat.

At the between-person level, the moment-to-moment lagged association from child defeat to maternal autonomy support during the puzzle task at T1 negatively predicted child externalizing problems at T2, above and beyond child externalizing problems at T1. That is, if the mother provided more autonomy support after the child showed more defeat during the puzzle task, this child had lower levels of externalizing problems six months later. The lagged association from maternal autonomy support to child defeat and the person means of maternal autonomy support and child defeat throughout the puzzle task did not significantly predict children's externalizing problems at T2. Child inhibitory control did not significantly predict children's externalizing problems at T2.

## The five-fold cross-validation

As shown in Figure 2, for the RDSEM without child inhibitory control as a predictor, the Mean Squared Error (MSE) across the five testing datasets was between .047 to .067 ($M$=.049), and the mean correlation between the predicted and observed child externalizing problems at T2 was between .58 to .63 ($M$=.61, .016<$p$<.029) for the 10 imputed datasets. For the RDSEM with child inhibitory control as a predictor, the MSE was between .050 to .057 ($M$=.053), and the mean correlation was between .60 to .63 ($M$=.61, .015<$p$<.018).

### *ASBIM*

As shown in Figure 2, for ASBIM model without child inhibitory control as a personalized individual feature, the Mean Squared Error (MSE) across the five testing datasets was between .042 to .055 ($M$=.046), and the mean correlation between the predicted and observed child externalizing problems at T2 was between .61 to .67 ($M$=.63, .010<$p$<.022) for the 10 imputed datasets. For ASBIM model including child inhibitory control as personalized information, the MSE was between .040 to .050 ($M$=.044), and the mean correlation was between .63 to .69 ($M$=.66, .003<$p$<.015). In addition, the $\gamma$ (see Equation 6) for both ASBIM without and with child inhibitory control across 10 imputed datasets are presented in Figure 3. The $\gamma$ is used to evaluate the importance of maternal autonomy support and child defeat in modeling mother-child behavior sequences. A higher value of $\gamma$ (i..e, $\gamma$>0.5) indicates a greater importance of maternal autonomy support, while a lower value (i..e, $\gamma$<0.5) suggests that child defeat plays a more significant role. As shown in Figure 3, the $\gamma$ for both models consistently exceeds 0.5 in all 10 imputed datasets. This consistent finding across multiple datasets underscores the significant role that maternal autonomy support plays in predicting the children's externalizing problems, suggesting that interventions aimed at improving maternal behaviors might be effective in mitigating the children's externalizing problems.

## Discussion

It is important to accurately predict children's future levels of externalizing problems, so that parents, teachers and practitioners can identify the children who are at the highest risk for developing externalizing problems and provide targeted prevention. Existing studies have shown that mothers providing support in response to children's negative emotions and dysregulated behaviors was associated with children's lower levels of externalizing problems.[10-12] Nevertheless, these studies focused on explaining the associations in the observed data, rather than examining the accuracy of prediction. In the current study, we examined the moment-to-moment fluctuations of maternal autonomy support and child defeat during a challenging puzzle task and to what extent children's externalizing problems six months later can be accurately predicted with these data. We examined prediction accuracy using five-fold cross-validation and compared the performance of a more theory-driven RDSEM and a more data-driven ASBIM model. We also examined whether including child inhibitory control in these models could further improve prediction accuracy. The results showed that in the RDSEM, for children whose mother displayed higher autonomy support after children expressed more defeat, these children had lower levels of externalizing problems six months later. Prediction accuracy of the RDSEM was good, with small MSEs and moderate correlations between the predicted values and the observed values of children's externalizing problems six months later. Prediction accuracy of the ASBIM model was higher than that of the RDSEM, especially after including children's inhibitory control in the model. Below we discuss the findings in more detail.

In consistent with previous studies,[10-12] we found that the lagged association between child defeat and maternal autonomy support predicted lower levels of children's externalizing problems six months later, after controlling for children's gender, their current levels of externalizing problems and the overall levels of maternal autonomy support and child defeat during the task. That is, for children whose mother provides higher autonomy support after children expressed higher defeat, they showed lower externalizing problems in the future. Mothers providing more autonomy support (e.g., asking questions to figure out the child's need, giving helpful instructions or tips, and providing encouragement) after children expressed defeat may provide children with helpful information regarding how to handle the challenging situation and the opportunity to practice self-regulation in a proactive way (see[19]). Such a parenting practice may facilitate children's internalization of regulatory skills helpful for reducing externalizing behaviors. Although it was posited that autonomy may be relatively less emphasized in collectivistic cultures,[55,56] in line with previous studies showing cultural similarity in the positive implications of maternal autonomy support for adolescents' emotional well-being,[43,57] we found that in a Chinese sample, mothers providing more autonomy support in response to more child defeat may be a protective factor against preschoolers' externalizing problems. This finding also corresponds with the results of ASBIM models showing that maternal autonomy support played a more important role than child defeat when predicting child externalizing behaviors, highlighting the importance of targeting maternal behaviors during parent-child interaction for preventing child externalizing problems.

Going beyond examining the associations between the observed data, we examined the prediction accuracy of the RDSEM using five-fold cross-validation. We found that the MSEs between the predicted and the observed values of child externalizing problems at T2 were smaller than the variance of child externalizing problems at T2 (.078) or the MSEs between child externalizing problems at T1 and child externalizing problems at T2 (.058 to .062, $M$=.061). In addition, the correlations between predicted and the observed values of child externalizing problems at T2 were moderate. These findings indicate that the RDSEM had good prediction accuracy and it provided more accurate prediction than directly using child externalizing problems at T1 as the predictor of child externalizing problems at T2.

Compared with the more theory-driven RDSEM, the more data-driven ASBIM model slightly improved prediction accuracy and showed smaller MSEs and higher correlations between the predicted and the observed values of children's externalizing problems at T2. As for the possible reason, unlike the more theory-driven RDSEM that rely on predefined assumptions and structures, the more data-driven ASBIM model can adopt deep learning to uncover complex patterns and nonlinear relationships that may not be immediately apparent. This ability allows ASBIM model to capture subtle mother-child interaction patterns within the data, leading to improved predictive accuracy. Yet, we acknowledge that a decrease of MSE by 6.12% and an increase of correlation by 3.28% were small. This may be partially due to our relatively small sample and the short length of observational task.

The ability of deep learning model to automatically mine high-order features and comprehensively consider all data may also help to explain why including child inhibitory control improved the prediction accuracy of the ASBIM model but not the RDSEM. In the RDSEM, child inhibitory control did not emerge as a significant predictor for child externalizing problems at T2 after controlling for the covariates and the lagged associations between fluctuations of maternal autonomy support and child defeat. We suspect that because children's externalizing problem were relatively stable across 6 months and our sample was not large, after controlling for child externalizing problems at T1, only especially strong predictors could reach significance. Given that child inhibitory control was not a significant predictor, including it in the RDSEM did not improve prediction accuracy. In contrast, in the ASBIM model, child inhibitory control was not only regarded as a predictor, but may also have interaction with fluctuations maternal autonomy support and child defeat. According to existing studies, children with lower inhibitory control may have more difficulty regulating their emotions and behaviors and thus may need parental support to a greater extent.[31-33] It is also possible that children with high inhibitory control may benefit from maternal autonomy support to a greater extent than those with low inhibitory control.[14] Thus, mothers providing more autonomy support after child expressing more defeat may have different predicting effect for children with different levels of inhibitory control. This may be the reason why including interactions between child inhibitory control and fluctuations of maternal autonomy support and child defeat help to improve the accuracy of predicting children's future externalizing problems.

We note several limitations of the current study and identify important future directions. First, the study had a relatively small sample. Even though a sample size of 101 aligns with various observational studies focusing on the dynamic mother-child interaction process in challenging problem-solving tasks during the preschool period,[23,24] this sample size remains relatively small for deep learning models. Despite implementing techniques like L2 regularization to combat overfitting, we are contemplating the incorporation of more cutting-edge approaches, such as contrastive learning, to tackle the challenge of limited sample sizes in forthcoming work. Additionally, we are exploring the possibility of enlarging the sample size or leveraging advanced data augmentation methods to expand the dataset in the future. Moreover, while deep learning enhances prediction performance, its black-box nature often results in limited interpretability of prediction outcomes. Future studies may consider leveraging the general language processing capabilities of large language models[58] to generate persuasive natural language explanations for the prediction processes of ASBIM model. Third, the current study was a short term longitudinal study. Given the stability of children's externalizing problems, children's externalizing problems at T1 greatly contributed to the prediction of children's externalizing problems at T2. It remains a question to what extent information about the dynamic process of mother-child interaction may contribute to long-term prediction of children's externalizing problems. Finally, the sample was recruited in a small city in China. To what extent the findings can be generalized to other socio-cultural contexts need to be examined. The finding that maternal autonomy supportive response to child defeat predicted children's lower levels of externalizing problems is consistent with patterns revealed in previous studies in the United States.[10-12] Thus, we expect that RDSEM and ASBIM based on mother-child interaction dynamics would have also good prediction accuracy for child externalizing problems in United States. Yet, this expectation need to be verified in future studies.

Despite these limitations, to our knowledge the current study was the first one to examine prediction accuracy for children's future externalizing problems using information about the dynamic process of mother-child interaction. Our study highlights the important role of maternal autonomy supportive response to child defeat. In addition, our study showed that the deep learning model helps to improve prediction accuracy, especially after including child inhibitory control in the model. Thus, the deep learning model can be a useful tool, when the goal is to accurately predict children's externalizing problems in order to identify the children who are at the greatest risk and to guide the targeted prevention.

# References

1. Eiden RD, Lessard J, Colder CR, Livingston J, Casey M, Leonard KE. Developmental cascade model for adolescent substance use from infancy to late adolescence. *Developmental Psychology*. 2016;52(10):1619-1633. doi:10.1037/dev0000199

2. Granic I, Patterson GR. Toward a comprehensive model of antisocial development: a dynamic systems approach. *Psychological Review*. 2006;113(1):101-131. doi:10.1037/0033-295X.113.1.101

3. Mesman J, Bongers IL, Koot HM. Preschool developmental pathways to preadolescent internalizing and externalizing problems. *Journal of Child Psychology and Psychiatry and Allied Disciplines*. 2001;42(5):679-689. doi:10.1017/S0021963001007351

4. Eisenberg N, Spinrad TL, Eggum ND. Emotion-related self-regulation and its relation to children's maladjustment. *Annual Review of Clinical Psychology*. 2010;6:495-525. doi:10.1146/annurev.clinpsy.121208.131208

5. Kopp, Claire B. Regulation of distress and negative emotions: A developmental view. *Developmental Psychology*. 1989;25(3):343. doi:10.1037/0012-1649.25.3.343

6. Beebe B, Steele M. How does microanalysis of mother-infant communication inform maternal

sensitivity and infant attachment?. *Attachment & Human Development*. 2013;15(5-6):583-602. doi:10.1080/14616734.2013.841050

7. Feldman R. Parent-infant synchrony and the construction of shared timing; physiological precursors, developmental outcomes, and risk conditions. *Journal of Child Psychology And Psychiatry*. 2007;48(3-4):329-354. doi:10.1111/j.1469-7610.2006.01701.x

8. Olson, Sheryl L., and Erika S. Lunkenheimer. Expanding concepts of self-regulation to social relationships: Transactional processes in the development of early behavioral adjustment. *A. Sameroff (Ed.), The Transactional Model of Development: How Children and Contexts Shape Each Other*. 2009;55-76. doi:10.1037/11877-004

9. Tronick EZ. Emotions and emotional communication in infants. *American Psychologist*. 1989;44(2):112-119. doi:10.1037//0003-066x.44.2.112

10. Lougheed JP, Hollenstein T, Lichtwarck-Aschoff A, Granic I. Maternal regulation of child affect in externalizing and typically-developing children. *Journal of Family Psychology*. 2015;29(1):10-19. doi:10.1037/a0038429

11. Ravindran N, Genaro BG, Cole PM. Parental Structuring in Response to Toddler Negative Emotion Predicts Children's Later Use of Distraction as a Self-Regulation Strategy for Waiting. *Child Development*. 2021;92(5):1969-1983. doi:10.1111/cdev.13563

12. Zhang X, Gatzke-Kopp LM, Fosco GM, Bierman KL. Parental support of self-regulation among children at risk for externalizing symptoms: Developmental trajectories of physiological regulation and behavioral adjustment. *Developmental Psychology*. 2020;56(3):528-540. doi:10.1037/dev0000794

13. Asparouhov, Tihomir, and Bengt Muthén. Comparison of models for the analysis of intensive longitudinal data. *Structural Equation Modeling: A Multidisciplinary Journal*. 2020;27(2);275-297. doi:10.1080/10705511.2019.1626733

14. Dong S, Dubas JS, Deković M, Wang Z. Goodness of fit in the Chinese context of socialization in the first three years. *Developmental Psychology*. 2022;58(10):1875-1886. doi:10.1037/dev0001409

15. Eisenberg N, Cumberland A, Spinrad TL, et al. The relations of regulation and emotionality to

children's externalizing and internalizing problem behavior. *Child Development*. 2001;72(4):1112-1134. doi:10.1111/1467-8624.00337

16. Lawler JM, Pitzen J, Aho KM, et al. Self-regulation and Psychopathology in Young Children. *Child Psychiatry & Human Development*. 2023;54(4):1167-1177. doi:10.1007/s10578-022-01322-x

17. Ravindran N, McElwain NL, Berry D, Kramer L. Dynamic fluctuations in maternal cardiac vagal tone moderate moment-to-moment associations between children's negative behavior and maternal emotional support. *Developmental Psychology*. 2022;58(2):286-296. doi:10.1037/dev0001299

18. Moorman EA, Pomerantz EM. The role of mothers' control in children's mastery orientation: a time frame analysis. *Journal of Family Psychology*. 2008;22(5):734-741. doi:10.1037/0893-3200.22.5.734

19. Grolnick, Wendy S., Alessandra J. Caruso, and Madeline R. Levitt. Parenting and children's self-regulation. *M. H. Bornstein (Ed.), Handbook of parenting: The practice of parenting (3rd ed.)*. 2019;34-64. doi:10.4324/9780429401695-2

20. Patterson GR, DeBaryshe BD, Ramsey E. A developmental perspective on antisocial behavior. *American Psychologist*. 1989;44(2):329-335. doi:10.1037//0003-066x.44.2.329

21. Cole PM, Teti LO, Zahn-Waxler C. Mutual emotion regulation and the stability of conduct problems between preschool and early school age. *Development And Psychopathology*. 2003;15(1):1-18.

22. Moed A, Dix T, Anderson ER, Greene SM. Expressing negative emotions to children: Mothers' aversion sensitivity and children's adjustment. *Journal of Family Psychology*. 2017;31(2):224-233. doi:10.1037/fam0000239

23. Lunkenheimer ES, Kemp CJ, Albrecht EC. Contingencies in Mother-Child Teaching Interactions and Behavior Regulation and Dysregulation in Early Childhood. *Social Development*. 2013;22(2):319-339. doi:10.1111/sode.12016

24. Lunkenheimer E, Ram N, Skowron EA, Yin P. Harsh parenting, child behavior problems, and the dynamic coupling of parents' and children's positive behaviors. *Journal of Family Psychology*. 2017;31(6):689-698. doi:10.1037/fam0000310

25. Neitzel C, Stright AD. Mothers' scaffolding of children's problem solving: establishing a foundation of academic self-regulatory competence. *Journal of Family Psychology*. 2003;17(1):147-159. doi:10.1037/0893-3200.17.1.147

26. Kim-Spoon J, Deater-Deckard K, Calkins SD, King-Casas B, Bell MA. Commonality between executive functioning and effortful control related to adjustment. *Journal of Applied Developmental Psychology*. 2019;60:47-55. doi:10.1016/j.appdev.2018.10.004

27. Garon N, Bryson SE, Smith IM. Executive function in preschoolers: a review using an integrative framework. *Psychological Bulletin*. 2008;134(1):31-60. doi:10.1037/0033-2909.134.1.31

28. Petersen IT, Hoyniak CP, McQuillan ME, Bates JE, Staples AD. Measuring the development of inhibitory control: The challenge of heterotypic continuity. *Developmental Review*. 2016;40:25-71. doi:10.1016/j.dr.2016.02.001

29. Kopp, Claire B. Antecedents of self-regulation: a developmental perspective. *Developmental Psychology*. 1982;18(2):199.doi:10.1037/0012-1649.18.2.199

30. Eisenberg N, Ma Y, Chang L, Zhou Q, West SG, Aiken L. Relations of effortful control, reactive undercontrol, and anger to Chinese children's adjustment. *Development And Psychopathology*. 2007;19(2):385-409. doi:10.1017/S0954579407070198

31. Reuben JD, Shaw DS, Neiderhiser JM, Natsuaki MN, Reiss D, Leve LD. Warm Parenting and Effortful Control in Toddlerhood: Independent and Interactive Predictors of School-Age Externalizing Behavior. *Journal of Abnormal Child Psychology*. 2016;44(6):1083-1096. doi:10.1007/s10802-015-0096-6

32. Van Aken C, Junger M, Verhoeven M, et al. The interactive effects of temperament and maternal parenting on toddlers' externalizing behaviours. *Infant and Child Development: An International Journal of Research and Practice*. 2007;16(5):553-572. doi:10.1002/icd.529

33. Yu J, Cheah CSL, Hart CH, Yang C. Child inhibitory control and maternal acculturation moderate effects of maternal parenting on Chinese American children's adjustment. *Developmental Psychology*. 2018;54(6):1111-1123. doi:10.1037/dev0000517

34. Satria, Andy, Opim Salim Sitompul, and Herman Mawengkang. 5-Fold Cross Validation on Supporting K-Nearest Neighbour Accuration of Making Consimilar Symptoms Disease

Classification. *2021 International Conference on Computer Science and Engineering*. 2021;1:1-5. doi:10.1109/IC2SE52832.2021.9792094

35. Zhu Y, Xi D, Song B, et al. Modeling users' behavior sequences with hierarchical explainable network for cross-domain fraud detection. *Proceedings of The Web Conference*. 2020;928-938. doi:10.1145/3366423.3380172

36. Han Z M, Huang C Q, Yu J H, et al. Identifying patterns of epistemic emotions with respect to interactions in massive online open courses using deep learning and social network analysis. Computers in Human Behavior. 2021;122:106843. doi:10.1016/j.chb.2021.106843

37. Sinclair J, Jang E E, Rudzicz F. Using machine learning to predict children's reading comprehension from linguistic features extracted from speech and writing. *Journal of Educational Psychology*. 2021;113(6):1088. doi:10.1037/edu0000658

38. Paolucci C, Giorgini F, Scheda R, et al. Early prediction of autism spectrum disorders through interaction analysis in home videos and explainable artificial intelligence. *Computers in Human Behavior*. 2023;148:107877. doi:10.1016/j.chb.2023.107877

39. Yan W, Yuan Y, Yang M, et al. Detecting the risk of bullying victimization among adolescents: A large-scale machine learning approach. *Computers in Human Behavior*. 2023;147:107817. doi:10.1016/j.chb.2023.107817

40. Waheed H, Hassan S U, Aljohani N R, et al. Predicting academic performance of students from VLE big data using deep learning models. *Computers in Human Behavior*. 2020;104:106189. doi:10.1016/j.chb.2019.106189

41. Chen X, Wang M, Huang H. Moment-to-moment within-person associations between maternal autonomy support and child defeat predicting child behavioral adjustment. *Journal of Family Psychology*. 2024;38(3):433-442. doi:10.1037/fam0001193

42. Ravindran N, McElwain NL. Dynamic coupling of maternal sensitivity and toddlers' responsive/assertive behaviors predicts children's behavior toward peers during the preschool years. *Developmental Psychology*. 2024;60(10):1801-1813. doi:10.1037/dev0001809

43. Cheung CS, Pomerantz EM, Wang M, Qu Y. Controlling and Autonomy-Supportive Parenting in the United States and China: Beyond Children's Reports. *Child Development*. 2016;87(6):1992-

2007. doi:10.1111/cdev.12567

44. NICHD Early Child Care Research Network. Child care and mother–child interaction in the first three years of life. *Developmental Psychology*. 1999;35(6):1399-1413. doi:10.1037/0012-1649.35.6.1399

45. Hallgren KA. Computing Inter-Rater Reliability for Observational Data: An Overview and Tutorial. *Tutorials in Quantitative Methods for Psychology*. 2012;8(1):23-34. doi:10.20982/tqmp.08.1.p023

46. Cole PM, Barrett KC, Zahn-Waxler C. Emotion displays in two-year-olds during mishaps. *Child Development*. 1992;63(2):314-324. doi:10.1111/j.1467-8624.1992.tb01629.x

47. Goodman R. The Strengths and Difficulties Questionnaire: a research note. *Journal of Child Psychology And Psychiatry*. 1997;38(5):581-586. doi:10.1111/j.1469-7610.1997.tb01545.x

48. Chen, Cheng-bin, and Jie Chen. Psychometric properties of Strengths and Difficulties Questionnaire-Chinese version: A systematic review. *Chinese Journal of Public Health*. 2017;33(4):685-688. doi:10.11847/zgggws2017-33-04-43

49. Putnam SP, Rothbart MK. Development of short and very short forms of the Children's Behavior Questionnaire. *Journal of Personality Assessment*. 2006;87(1):102-112. doi:10.1207/s15327752jpa8701_09

50. Xu Y, Farver JA, Zhang Z. Temperament, harsh and indulgent parenting, and Chinese children's proactive and reactive aggression. *Child Development*. 2009;80(1):244-258. doi:10.1111/j.1467-8624.2008.01257.x

51. Guo H, Chen B, Tang R, et al. An embedding learning framework for numerical features in ctr prediction. *Proceedings of the 27th ACM SIGKDD Conference on Knowledge Discovery & Data Mining*. 2021;2910-2918. doi:10.1145/3447548.3467077

52. Zhang Y, Liu C, Liu M, et al. Attention is all you need: utilizing attention in AI-enabled drug discovery. *Briefings in Bioinformatics*. 2023;25(1):bbad467. doi:10.1093/bib/bbad467

53. Wang D, Xu D, Yu D, et al. Time-aware sequence model for next-item recommendation[J]. Applied Intelligence. 2021;51:906-920. doi:10.1007/s10489-020-01820-2

54. Yi D, Ahn J, Ji S. An effective optimization method for machine learning based on ADAM.

*Applied Sciences*. 2020;10(3):1073. doi:10.3390/app10031073

55. Markus H R, Kitayama S. Culture and the self: Implications for cognition, emotion, and motivation. *College Student Development and Academic Life*. 2014;264-293. doi:10.1037/0033-295X.98.2.224

56. Trommsdorff G, Cole P M, Heikamp T. Cultural variations in mothers' intuitive theories: A preliminary report on interviewing mothers from five nations about their socialization of children's emotions. *Global Studies of Childhood*. 2012;2(2):158-169. doi:10.2304/gsch.2012.2.2.158

57. Xiong Y, Qin L, Wang Q, Wang M, Pomerantz EM. Reexamining the cultural specificity of controlling and autonomy-supportive parenting in the United States and China with a within-individual analytic approach. *Developmental Psychology*. 2022;58(5):935-949. doi:10.1037/dev0001329

58. Min B, Ross H, Sulem E, et al. Recent advances in natural language processing via large pre-trained language models: A survey. *ACM Computing Surveys*. 2023;56(2):1-40. doi:10.1145/3605943

**Table 1** Descriptive statistics and bivariate correlations for the study variables (*N*=101).

| Variable | | 1. | 2. | 3. | 4. | 5. |
|---|---|---|---|---|---|---|
| 1. Maternal autonomy support T1 | | --- | .15* | --- | --- | --- |
| 2. Child defeat T1 | | .12 | --- | --- | --- | --- |
| 3. Externalizing problems T1 | | .10 | .10 | --- | --- | --- |
| 4. Externalizing problems T2 | | .10 | -.03 | .65*** | --- | --- |
| 5. Inhibitory control T1 | | -.05 | -.24* | -.49*** | -.45*** | --- |
| | Mean (SD) | .58 (.29) | .21 (.17) | .69 (.28) | .62 (.28) | 4.78 (.60) |
| | Range | 0-1.45 | 0-.70 | .10-1.35 | .05-1.30 | 3.38-6.50 |
| | ICC | 17.42% | 23.02% | --- | --- | --- |

Note, *$p$<.05, ***$p$<.001. Between-person correlations are below the diagonal and within-person correlations are above the diagonal

**Table 2** Parameter estimates for the main model with the full sample (*N*=101).

| Parameters | Externalizing problems T2 | | Externalizing problems T2 | |
|---|---|---|---|---|
| | *Est. (SD$_{posterior}$)* | 95% CI | *Est. (SD$_{posterior}$)* | 95% CI |
| **Intercept** | .30 (.10)* | [.12, .50] | .77 (.47) | [-.16, 1.71] |
| **Within-Person Level** | | | | |
| Maut$_t$ → Cdef$_{t+1}$ [Mother-to-Child Association] | .17 (.06)* | [.06, .29] | .16 (.06)* | [.05, .29] |
| Cdef$_t$ → Maut$_{t+1}$ [Child-to-Mother Association] | .11 (.06) | [-.00, .21] | .10 (.05) | [-.01, .21] |
| Covariance: Maut$_t$ with Cdef$_t$ | .16 (.05)* | [.07, .26] | .16 (.05)* | [.06, .25] |
| Residuals: Maut$_t$ → Maut$_{t+1}$ | .10 (.04)* | [.02, .18] | .10 (.04)* | [.02, .18] |
| Residuals: Cdef$_t$ → Cdef$_{t+1}$ | .21 (.07)* | [.08, .33] | .21 (.07)* | [.08, .34] |
| $t$ → Maut$_t$ | -.00 (.01) | [-.03, .02] | -.00 (.01) | [-.03, .02] |
| $t$ → Cdef$_t$ | .02 (.02) | [-.01, .05] | 02 (.02) | [-.01, .05] |
| **Between-Person Level** | | | | |
| [Mother-to-Child Association] → Child outcome T2 | .10 (.18) | [-.24, .46] | .02 (.19) | [-.37, .39] |
| [Child-to-Mother Association] → Child outcome T2 | -.44 (.22)* | [-.91, -.04] | -.53 (.22)* | [-.98, -.12] |
| Maut → Child outcome T2 | .11 (.15) | [-.18, .43] | .07 (.16) | [-.24, .41] |
| Cdef → Child outcome T2 | .11 (.18) | [-.24, .50] | .08 (.19) | [-.29, .48] |
| Child outcome T1 → Child outcome T2 | .54 (.12)* | [.28, .78] | .48 (.13)* | [.21, .72] |
| Child gender → Child outcome T2 | -.05 (.05) | [-.14, .04] | -.05 (.05) | [-.14, .04] |
| Child inhibitory control → Child outcome T2 | --- | --- | -.08 (.09) | [-.26, .10] |

Note, Maut=maternal autonomy support. Cdef=child defeat. *t*=interval. child gender 0=boys, 1=girls.

*=significant at α.05. Variances of the within-person level parameters and their covariances with each other and with the between-person predictors were estimated but not shown

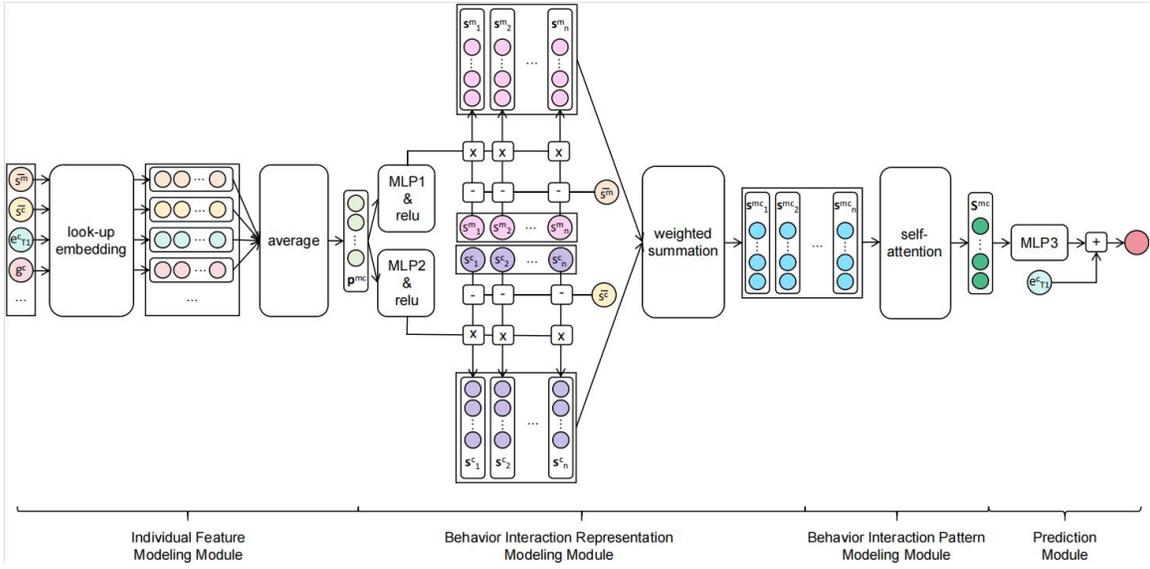

**Figure 1** Framework of the Attention-based Sequential Behavior Interaction Modeling (ASBIM) model.

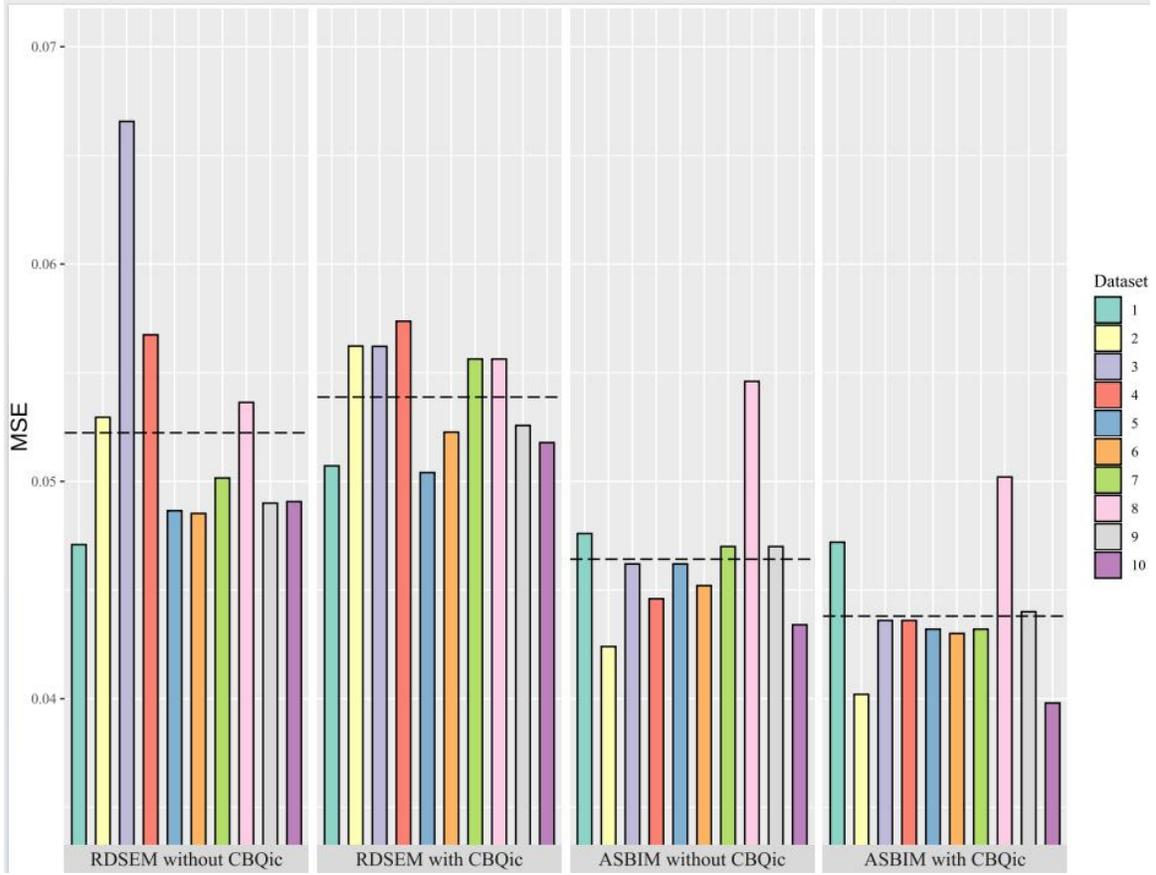

**Figure 2(a)** Mean Square Error (MSE) between the Predicted and Observed Values of T2 Externalizing Problems for the RDSEM and the ASBIM models with and without Child Inhibitory Control

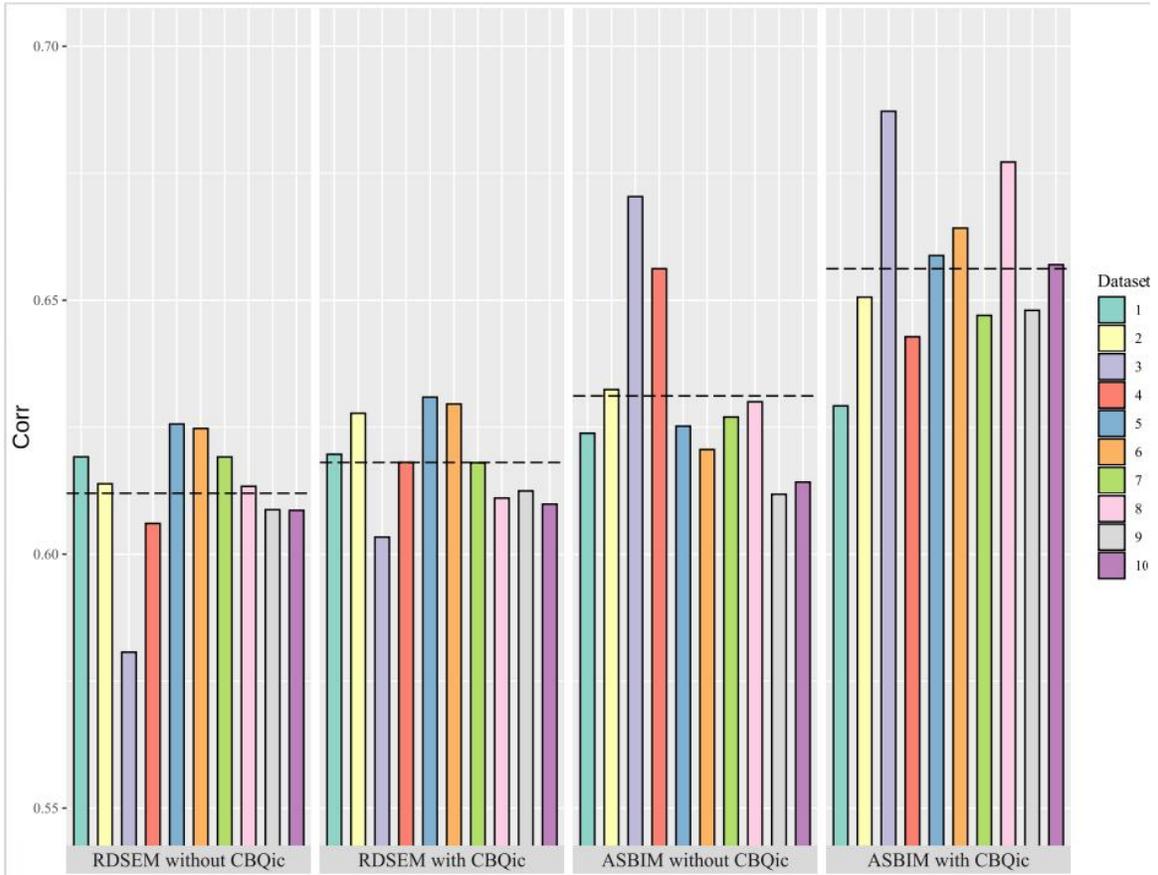

**Figure 2(b)** Correlation between the Predicted and Observed Values of T2 Externalizing Problems for the RDSEM and the ASBIM models with and without Child Inhibitory Control

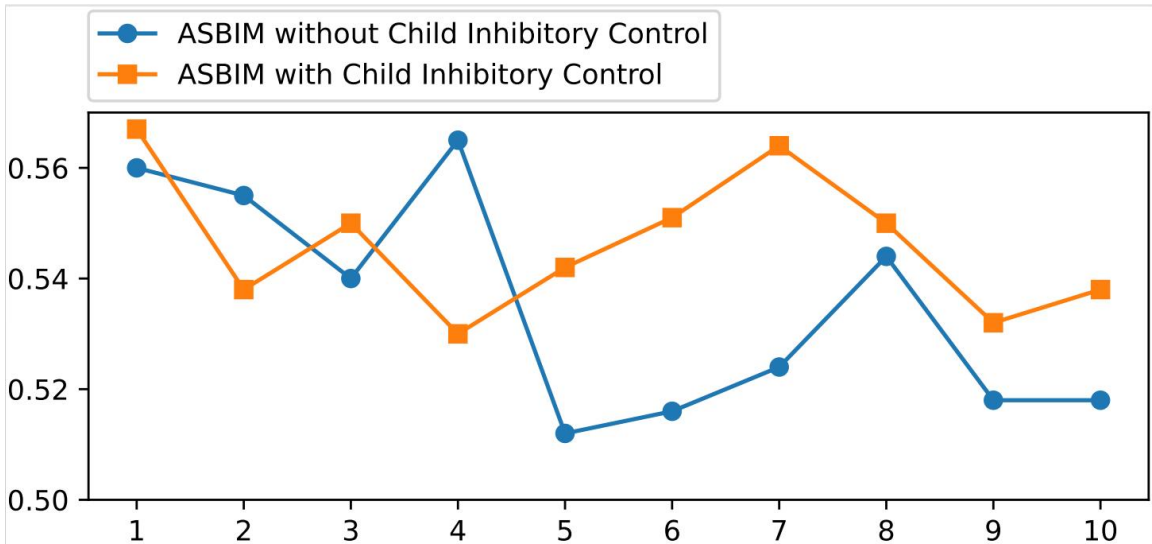
**Figure 3** The γ for both ASBIM model and ASBIM+Child Inhibitory Control model across 10 imputed datasets.